# European Language Grid: An Overview

**Georg Rehm[1], Maria Berger[1], Ela Elsholz[1], Stefanie Hegele[1], Florian Kintzel[1], Katrin Marheinecke[1], Stelios Piperidis[2], Miltos Deligiannis[2], Dimitrios Galanis[2], Katerina Gkirtzou[2], Penny Labropoulou[2], Kalina Bontcheva[3], David Jones[3], Ian Roberts[3], Jan Hajič[4], Jana Hamrlová[4], Lukáš Kačena[4], Khalid Choukri[5], Victoria Arranz[5], Andrejs Vasiļjevs[6], Orians Anvari[6], Andis Lagzdiņš[6], Jūlija Meļņika[6], Gerhard Backfried[7], Erinç Dikici[7], Miroslav Janosik[7], Katja Prinz[7], Christoph Prinz[7], Severin Stampler[7], Dorothea Thomas-Aniola[7], José Manuel Gómez Pérez[8], Andres Garcia Silva[8], Cristian Berrío[8], Ulrich Germann[9], Steve Renals[9], Ondrej Klejch[9]**

[1] DFKI GmbH, Germany • [2] ILSP/Athena RC, Greece • [3] University of Sheffield, UK • [4] Charles University, Czech Republic • [5] Evaluations and Language Resources Distribution Agency (ELDA), France • [6] Tilde, Latvia • [7] SAIL LABS Technology GmbH, Austria • [8] Expert System Iberia SL, Spain • [9] University of Edinburgh, UK

Corresponding author: Georg Rehm – georg.rehm@dfki.de

## Abstract

With 24 official EU and many additional languages, multilingualism in Europe and an inclusive Digital Single Market can only be enabled through Language Technologies (LTs). European LT business is dominated by hundreds of SMEs and a few large players. Many are world-class, with technologies that outperform the global players. However, European LT business is also fragmented – by nation states, languages, verticals and sectors –, significantly holding back its impact. The European Language Grid (ELG) project addresses this fragmentation by establishing the ELG as the primary platform for LT in Europe. The ELG is a scalable cloud platform, providing, in an easy-to-integrate way, access to hundreds of commercial and non-commercial LTs for all European languages, including running tools and services as well as data sets and resources. Once fully operational, it will enable the commercial and non-commercial European LT community to deposit and upload their technologies and data sets into the ELG, to deploy them through the grid, and to connect with other resources. The ELG will boost the Multilingual Digital Single Market towards a thriving European LT community, creating new jobs and opportunities. Furthermore, the ELG project organises two open calls for up to 20 pilot projects. It also sets up 32 National Competence Centres (NCCs) and the European LT Council (LTC) for outreach and coordination purposes.

## 1. Introduction

With 24 official EU languages and many additional ones, multilingualism, cross-lingual and cross-cultural communication in Europe as well as an inclusive EU Digital Single Market can only be enabled and firmly established through Language Technologies (LTs) (Rehm, 2016). The European LT industry is dominated by hundreds of SMEs and a few large players. Many are world-class, with technologies that outperform the global players. However, European LT business is also fragmented – by nation states, languages, domains and sectors (Vasiljevs et al., 2019) –, significantly holding back its impact. In addition, many European languages are severely under-resourced and, thus, in danger of digital language exinction (Rehm and Uszkoreit, 2012; Kornai, 2013; Rehm et al., 2014; Rehm et al., 2016a), which is why there is an enormous need for a European LT platform as a unifying umbrella (Rehm and Uszkoreit, 2013; Rehm et al., 2016b; STOA, 2017; Rehm, 2017; Rehm and Hegele, 2018; European Parliament, 2018).

The project European Language Grid (ELG; 2019-2021) addresses this fragmentation by establishing the ELG as the primary platform and marketplace for the European LT community, both industry and research.[1] The ELG is developed to be a scalable cloud platform, providing, in an easy-to-integrate way, access to hundreds of commercial and non-commercial LTs for all European languages, including running tools and services as well as data sets and resources. Once fully operational, it will enable the commercial and non-commercial European LT community to upload their technologies and data sets into the ELG in an easy and efficient way, to deploy them through the grid, and to connect with other resources. The ELG will boost the Multilingual Digital Single Market towards a thriving European LT community, creating new jobs and opportunities, also addressing the threat of digital language extinction.

## 2. Approach and Methodology

The European LT community has been demanding a dedicated LT platform for years (Section 1). The ELG project, whose platform is supposed to fill this gap, has various objectives. Its ambition is to establish the ELG as the primary platform for industry-relevant LT in Europe, bringing together and uniting a network of European experts and concentrating on *commercial* and *non-commercial LTs* (i. e., LTs with a high Technology Readiness Level, TRL), both *functional* (processing and generation for written and spoken language) and *non-functional* (corpora, lexicons, data sets etc.). A closely related goal is to establish the ELG as the primary market place for the fragmented European LT landscape (Vasiljevs et al., 2019) to connect demand and supply, strengthening Europe's position in this field. The platform is meant to enable the whole European LT community to upload their services and data sets, to deploy them and to connect with, and make use of those resources made available by others (taking into account IPR and licenses, and including payment and billing options, esp. with regard to commercial resources). *The ELG is meant to be a shared platform for the whole European LT community*, enabling not only LT provider companies to grow and benefit from scaling up but also companies who want to integrate LT into their products or services. The ELG consortium consists of nine partners from research and industry (see the affiliations of the co-authors of this article) that either lead or participate in important community initiatives or have good relationships with other relevant initiatives.

---
[1] https://www.european-language-grid.eu

The project is structured into three areas that relate to the *Grid Platform*, the *Grid Content* and the *Grid Community*. The *Grid Platform* takes care of setting up and developing the base infrastructure. The platform is built with robust, scalable, reliable and widely used open source technologies that are constantly developed further, enabling it to scale with the growing demand and supply. The ELG catalogue contains records of all resources (including services, data sets etc.) as well as records of LT companies, research organisations, projects, service and application types, languages etc. This is where the first area overlaps with the second, i. e., *Grid Content*, which is the actual content of the ELG in terms of processing or generation services, tools, data sets, corpora, language resources etc. We distinguish between functional content (running services that can be uploaded into the ELG and integrated into other systems) and non-functional content (corpora, data sets etc.). Functional LT services are realised by containerising and ingesting them into the ELG (including metadata). Our goal is to make this process as easy and efficient as possible for commercial and non-commercial LT providers. These are two of the main groups of the *Grid Community*, i. e., all stakeholders of the ELG, which also include companies that want to purchase or integrate LT, public administrations, NGOs etc. The project collaborates with the six research projects funded through ICT-29-2018 subtopic b), META-NET, LT Innovate and also with projects such as AI4EU (Rehm et al., 2020b). In addition, ELG established a network of 32 National Competence Centres (NCCs) in 32 European countries. The NCCs act as national bridges to identify content and to interest relevant stakeholders in participating in the ELG initiative. On a broader level, ELG is establishing the European LT Council as a pan-European body, in which LT-related matters can be coordinated. Finally, in 2020 ELG publishes two open calls through which a total of 15-20 pilot projects will be financially supported. These will extend the ELG's catalogue with relevant services or data sets and realise innoative applications based on the ELG, demonstrating the usefulness of the platform.

## 3. The European Language Grid

In the following, we describe the technical architecture of the ELG cloud platform (Section 3.1), including the catalogue and metadata schema (Section 3.2) as well as the graphical user interface (Section 3.3). Section 3.4 provides more details on the functional services available in the ELG including the generic API approach. The data sets and language resources are discussed in Section 3.5, followed by a short description of the ELG community and other stakeholders (Section 3.6). Section 3.7 provides an overview of the two open calls for pilot projects.

### 3.1. Technical Architecture

ELG is a scalable platform with an interactive web user interface and corresponding backend components and REST APIs. It offers access to various kinds of resources such as corpora and data sets as well as functional LT services, i. e., existing LT tools that have been containerised and wrapped with the ELG LT Service API.[2] ELG's integrated functional services can be used through APIs or through the web interface. The architecture is separated into three layers (Figure 1), i. e., the *base infrastructure* (Kintzel et al., 2019; Moritz et al., 2019), the *platform backend* (Piperidis et al., 2019; Labropoulou et al., 2019) and the *platform frontend* (Melnika et al., 2019a; Melnika et al., 2019b).

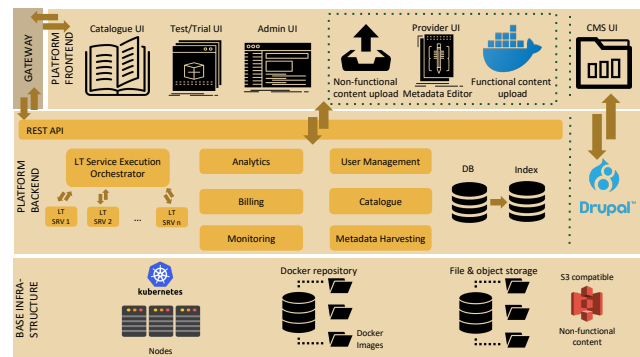

Figure 1: Technical architecture of the ELG

The *base infrastructure* is operated on a Kubernetes[3] cluster in the data centre of Syseleven, a European provider located in Berlin, Germany. All infrastructural components of the three layers run as Docker containers in this cluster. They are built with robust, scalable, reliable and widely used technologies/frameworks that are constantly developed further, e. g., Django, Spring Boot, AngularJS and ReactJS.

The *platform backend* contains the ELG catalogue, i. e., the central list of metadata records of functional services, non-functional resources (e. g., data sets) but also the entries of organisations (e. g., companies, universities, research centres) and other stakeholders, as well as service types, languages and other types of information (Section 3.2). Once ELG is fully deployed, stakeholders will be able to register themselves in this catalogue, ensuring increased reach and visibility. Users can filter and search for organisations, services, data sets and more, by languages, service types, domains, and countries. The catalogue is implemented using Django, its functionalities are offered via REST services which consume and produce JSON messages. The catalogue uses a PostgreSQL database for storing metadata records and an ElasticSearch engine. The platform backend layer also includes the LT Service Execution Server that offers a common REST API for calling integrated functional services. It orchestrates service execution by calling the LT service, also handling failures, timeouts etc. Other backend components are work in progress, e. g., for billing, monitoring the infrastructural components, editing metadata records as well as for user management.

---

[2]https://gitlab.com/european-language-grid/platform/
[3]https://kubernetes.io

The *platform frontend* layer consists of UIs for the different types of ELG users, e. g., LT providers, potential buyers and ELG system administrators (Section 3.3). These include Catalogue UIs (browse, faceted or text-based search, edit metadata), test/trial UIs for functional services, provider UIs for uploading/registering functional services etc. They are implemented using ReactJS and Django by exploiting the catalogue backend services, e. g., a resource's metadata record is returned as a JSON object and rendered as HTML. The frontend also includes a Drupal-based CMS that contains ELG-related content and information (Section 3.3).

All publicly available ELG services (e. g., UIs, LT Service Execution API) are served via an nginx webserver which also runs as a container. The use of containers and Kubernetes facilitates the platform's deployment and portability. One of the key concepts of the architecture is the use of containers to encapsulate all components, settings and libraries of an individual LT service in one self-contained unit. Docker is currently the most widely used technology for containerisation. In our context this means that, for individual LT services, Docker images can be built locally by their respective providers and ingested into the ELG, where they can be started, terminated and scaled out on demand. Containers can also be replaced easily by their providers. The containerisation of LT services solves to a great extent interoperability and deployment issues, which is particularly useful for ELG since heterogeneous technologies are used for LT development, e. g., different programming languages, operating systems, frameworks and libraries.

Kubernetes is used for container orchestration in all ELG layers. For the LT service containers in particular, the platform makes use of knative[4], a layer on top of Kubernetes that handles auto-scaling to match demand, including scaling down to zero when there is no demand and back up when requests begin again. The platform also takes advantage of facilities in Kubernetes and knative to monitor running containers, to detect and proactively terminate any that become unresponsive or to restart those that have crashed.

One of the most important goals of ELG is a) to enable commercial or non-commercial providers to adapt their LT services so that they can be integrated efficiently, b) to make the ingestion of their containerised services into the ELG, i. e., the upload and description with metadata, as simple as possible. Currently, this integration of a service consists of six steps: (1) adapt the service to fit the ELG API; (2) create a Docker image for the service; (3) push the Docker image into a registry (e. g., ELG Gitlab); (4) request, from the ELG administrators, a Kubernetes namespace[5], in case of a proprietary service with restricted access; (5) deploy the service by creating the respective Kubernetes config file; (6) add the service to the ELG catalogue by contacting the ELG administrators and providing the metadata.[6] For some of the more than 100 services currently in the ELG, this process took a few days, for others, only a few hours. Our medium to long term goal is to bring this effort down to a minimum, at least for the most common cases (e. g., a Python based LT service created with a well-known machine learning framework), by providing Docker templates.

### 3.2. Catalogue Structure – Metadata Schema

The ELG catalogue contains all entities of interest to users (Section 3.6), appropriately indexed and described so that they can easily search, find and select the resources that meet their requirements and deploy them, as well as visualise the LT domain activities, stakeholders and resources with specific criteria (e. g., service type, language, etc.). All entities are described in compliance with the ELG-SHARE metadata schema (Labropoulou et al., 2019; Labropoulou et al., 2020).[7] The schema builds upon, consolidates and updates previous activities, especially the META-SHARE schema and its profiles (Gavrilidou et al., 2012; Piperidis et al., 2018; Labropoulou et al., 2018), taking into account the ELG user requirements (Melnika et al., 2019a), recent developments in the (meta)data domain (e. g., FAIR[8], data and software citation recommendations[9], Open Science movement, etc.), and the need for establishing a common pool of resources through exchange mechanisms with collaborating projects and initiatives (Rehm et al., 2020c), cf. Section 3.6. The schema caters for the description of the ELG core entities (Figure 2), i. e., Language Technologies (*tools/services*), including functional services and non-functional ones (e. g., downloadable tools, software code, etc.), and Data Language Resources, comprising *data sets* (corpora), *language descriptions* (i. e., models and computational grammars) and *lexical/conceptual resources* (e. g., gazetteers, ontologies, term lists, etc.). It also provides for entities involved in their production and usage and, in general, LT activities, namely *actors* (*organizations*, *groups* and *persons*), *documents* (e. g., user manuals, publications, etc.), *projects* and *licences/terms of use*. The schema defines metadata elements for each entity type, capturing properties of LRTs throughout all stages of their lifecycle from production to usage, properties of the related entities with regard to their LT activities, and relations between them, resulting in a schema which is rich in information and can provide a global view of the LT landscape. However, only a subset of carefully selected elements is mandatory. Thus, minimal metadata records can be imported, both from resource providers, and from other sources through harvesting and conversion APIs (see Sections 3.4 and 3.5), gradually enriched through (semi-)automatic processes and curated by persons who rightfully claim them.

### 3.3. User Interface

To identify the user scenarios and requirements for designing the interface, we defined the main groups of ELG users: (1) *Content providers* – companies, research organisations or public institutions with tools, services, or data that can be provided through the ELG; (2) *Developers and integrators* – companies and research institutions interested in using LT services, tools, or data in their applications; (3) *Information seekers* – users interested in information about LT, data or events; (4) *Information providers* – organisations or individuals who wish to provide information about events, trainings etc.; (5) *Casual visitors* without any professional connection to LTs; (6) *ELG platform administrators*.

---

[4]https://knative.dev

[5]A virtual sub-cluster, which can be used to restrict access to the respective containers that run within it.

[6]Each LT service is tested for conformity to ELG specifications before it is published to the catalogue, i. e., we test that it can be called from the LT Service orchestrator and return an appropriate JSON response that can also be rendered from the try out GUIs.

[7]The version for ELG Release 1 with documentation and examples is available at https://gitlab.com/european-language-grid/platform/ELG-SHARE-schema under a CC-BY-4.0 licence.

[8]https://www.force11.org/group/fairgroup/fairprinciples

[9]https://www.force11.org/datacitationprinciples

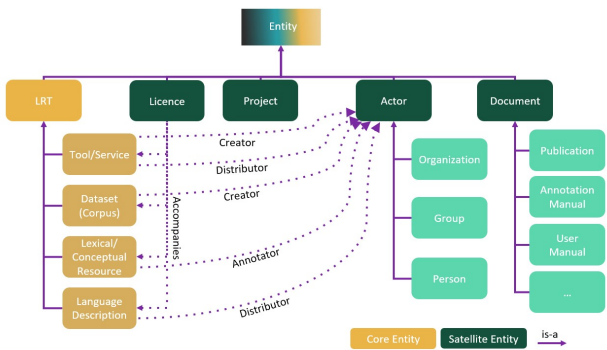

Figure 2: Entities covered by the ELG metadata schema

We carried out a user survey and interviews to elicit and prioritise the user requirements, 158 respondents participated in the survey (50 responses were incomplete and, thus, excluded from the analysis). 67% of the respondents were from academia/research, 24% from industry, and the rest from public administrations and NGOs. Some of the survey findings are: 70% of respondents are interested in providing software as a container. Tool, service, data provider respondents selected an easy and quick process (97%), secure access (74%), the possibility to provide metadata (73%), and usage statistics (68%) as the most important features. When searching for LT software or services, important criteria are language coverage (62%), license and access conditions (59%) and availability of open source code (56%).

The ELG corporate identity is based on various elements (e. g., logo, colours, typography) developed in collaboration with a graphics designer. The Google Material Design[10] framework is used to integrate the ELG identity and to develop the ELG web design components in the ELG CMS front-end and the Catalogue UI application.

Angular and Typescript – a superset of JavaScript – are used for developing the ELG CMS front-end (cf. Figure 3 in the appendix), while for the Catalogue UI (browse, faceted or text-based search, edit metadata, download/run services and data, etc.) we use the React library. Both Angular and React Material components are implemented as an adjustable theme that can be tuned to the designer's specifications; the website design for both applications is based on the Single Page Application (SPA) principle. Rather than reloading each page in its entirety, SPAs load content dynamically, which improves page loading speed significantly because only parts of a page have to be updated. The ELG Angular application uses ngx-cache for in-app cache management of local data. For SEO optimisation ngx-meta is used, which enables the creation of SEO-friendly URLs and generates title, meta tags, and Open Graph tags for social media sharing.[11] The Catalogue UI makes use of the React-Helmet library for the creation of the meta tags elements. To facilitate the (future) localisation of the ELG interface into different languages we use ngx-translate.[12] JSON files contain the translatable content as key-value pairs, which can be easily translated into any language as needed. Currently, both web applications (CMS and Catalogue UI) use client-side-rendering, i. e., they return a single HTML file to the browser and the rest of the application comes as a set of Javascript files. In the case of the Catalogue UI, the Javascript files use the backend catalogue REST services (which produce and consume JSON). User authorization is ensured by adding a JSON Web Token (JWT) to data requests, where the user identity data is encoded and sent as an encrypted JSON object. ELG follows the W3C Web Content Accessibility Guidelines.[13]

To enable the flexible management of content within the ELG website, we integrated Drupal (version 8). As the ELG CMS front-end is a monolithic SPA, the CMS does not have a dedicated public front-end. Instead, Drupal serves different menus and page contents using REST services and JSON-HAL. The built-in Drupal front-end is protected by a password, and, after successful authentication, the user is redirected to the CMS environment and authorised for specific operations based on her/his role. Authored content is served using REST and JSON-HAL. The Angular SPA front-end application transforms the data into a layout and portal design that is specific for the user scenario. Static files (PDFs, photos, graphics, etc.) are provided directly from the CMS. All front-end components including the CMS are containerised and delivered to be ready for deployment through Kubernetes (Section 3.1).

### 3.4. Functional Services

The European LT market is very broad and varied, with many different providers of many different classes of services and tools, exposed as many different APIs and data formats. One of ELG's primary goals is to attempt to bring more order to this varied landscape by identifying classes of related services and providing generic APIs for each class. From the outset the project has identified a number of broad classes that share much in common:

*Machine Translation (MT)*: services that take text in one language and translate it into text in another language, possibly with additional metadata associated with each segment (sentence, phrase, etc.). This class can include (seemingly unrelated) services such as summarisation, where the summary can be viewed as a "translation" of the original text.

*Information Extraction (IE)*: services that take text and annotate it with metadata on specific segments. This class can cover a wide variety of services from basic NER through to complex sentiment analysis and domain-specific tools.

*Automatic Speech Recognition (ASR)*: services that take audio as input and produce text (e. g., a transcription) as output, possibly with metadata associated with each segment. Other clusters are emerging as the project considers more services for integration, for example text-to-speech, text classification, alignment, and translation quality estimation. An aspiration for the platform is to provide services of all classes for all official EU languages and for other EU and non-EU languages that are of strategic interest within the EU. As a first step, the project has conducted a survey of all LT services available among the ELG partners, as well as third-party open source tools that can fill the obvious gaps. Table 1 shows the total numbers of services of each type available within the consortium, see Tables 5 and 6 in the appendix. For the purpose of this analysis, languages have been divided into four groups: (A) EU official languages; (B) other EU languages without official status, plus languages from candidate countries and free trade partners; (C) languages spoken by immigrants or important trade and political partners; (D) languages that do not fit (A), (B), (C).

---

[10]https://material.io/design
[11]https://github.com/fulls1z3/ngx-meta
[12]https://github.com/ngx-translate/core
[13]https://www.w3.org/WAI/standards-guidelines/wcag

|                              | A   | B  | C   | D  |
|------------------------------|-----|----|-----|----|
| **ASR**                      | 12  | 3  | 9   |    |
|   Speech recognition | 12 | 3  | 9   |    |
| **IE & Text Analysis**       | 311 | 50 | 119 | 12 |
|   Terminology markup | 24 | 1  | 1   |    |
|   Part of speech tagging | 24 | 7 | 13 |    |
|   Tokenization     | 24  | 7  | 13  |    |
|   Dependency parsing | 24 | 7  | 13  |    |
|   Term recognition | 24  | 1  | 1   |    |
|   Lemmatisation    | 24  | 7  | 12  |    |
|   Terminology search | 24 | 1  | 1   |    |
|   Morphological analyser | 24 | 7 | 13 |    |
|   Language identification | 22 | 4 | 14 | 12 |
|   Named entity recognition | 18 | 5 | 11 |    |
|   Keyword extraction | 9  | 3  | 9   |    |
|   Sentiment analysis | 8  |    | 4   |    |
|   Summarization    | 7   |    | 5   |    |
|   Polarity detection | 7  |    | 4   |    |
|   Key phrase extraction | 7 |    | 5   |    |
|   Categorization   | 5   |    |     |    |
|   Sentence splitting | 4 |    |     |    |
|   NER disambiguation | 4 |    |     |    |
|   Textual entailment | 3 |    |     |    |
|   Information extraction | 3 |  |     |    |
|   Entity linking   | 2   |    |     |    |
|   Time annotation  | 2   |    |     |    |
|   Proofing tools   | 2   |    |     |    |
|   Date detection   | 2   |    |     |    |
|   Number annotation | 2  |    |     |    |
|   Relation extraction | 1 |    |     |    |
|   Negation detection | 1 |    |     |    |
|   Text extraction  | 1   |    |     |    |
|   Number normalisation | 1 |   |     |    |
|   Measurement normalisation | 1 | | |    |
|   Opinion mining   | 1   |    |     |    |
|   Noun phrase extraction | 1 |  |     |    |
|   Parsing          | 1   |    |     |    |
|   Measurement annotation | 1 |  |     |    |
| **MT** (Source ↓ / Target →) | 67  | 2  | 7   | 1  |
|   A                | 58  | 2  | 7   | 1  |
|   B                | 2   |    |     |    |
|   C                | 7   |    |     |    |
| **Text to Speech**           | 7   | 1  | 1   | 2  |
|   Text to speech   | 7   | 1  | 1   | 2  |
| **Other**                    | 27  | 2  |     |    |
|   Web crawler      | 24  | 2  |     |    |
|   Text indexing    | 2   |    |     |    |
|   Pipeline         | 1   |    |     |    |
| **Total general**            | 424 | 58 | 136 | 15 |

Table 1: Services provided by the ELG partners with support for languages in the four categories (A, B, C, D)

The first prototype (Oct. 2019) included seven IE and text analysis tools plus a dependency parser supporting 60 languages, five ASR services (one supporting three languages and another supporting two), 14 MT services (six languages into English, English into eight other languages) and text-to-speech in four languages. The current version, Release 1 alpha (launched in early March 2020), is populated with 133 IE and text analysis services (incl. a multilingual dependency parser supporting 60 languages), 24 MT, nine ASR, four TTS and two text categorization services.

We aim to make it as simple as possible for LT providers to integrate their services, but in a way that avoids the proliferation of incompatible APIs for the same task, allowing users to access the widest range of services without being locked in to a single vendor. We have defined a generic API for each class of services with flexible and reusable JSON formats for request and response payloads. Service providers who want to integrate their services into the ELG only need to provide a Docker image that presents an HTTP endpoint that can receive requests and return responses in the specified format. The ELG platform will receive requests from external users, handle user authentication, authorization, billing, etc., validate the user input and make the required calls to the containerised LT services as per the ELG specification, receive and validate the responses from the containers and finally respond to the caller as required.[14]

In the current version, the public-facing APIs mirror very closely the internal API between the platform and the containerised services but this approach will make it possible for ELG to offer any number of public-facing endpoints matching popular third-party API specifications, map these to the internal ELG format and thus present *all* services of that class under the same public APIs. Future developments may also include other styles of interaction such as batch processing of resources hosted in the catalogue, without requiring changes to the underlying container images.

The required LT service provider APIs are documented in the OpenAPI format, and there are libraries available for most popular programming languages to take OpenAPI definitions and produce skeleton code to serve as the basis for the service implementation. ELG will provide custom helper libraries for the most common web service implementation libraries to further reduce the amount of boilerplate code required. As an example, for the first prototype, ELG consortium members were able to integrate a variety of their own open source and commercial services, written in disparate programming languages (Java/Spring, .NET, Python) with just a few days work in the first iteration, falling to a few hours once developers became more familiar with the infrastructure and required formats.

---

[14] These requests are received and handled by the LT Service Execution Server (Section 3.1).

The composition of individual services offered by ELG directly or other cloud platforms is not addressed by ELG itself. However, we experiment with workflow composition and platform interoperability in other contexts (Rehm et al., 2020a; Rehm et al., 2020b; Moreno-Schneider et al., 2020a; Moreno-Schneider et al., 2020b) and plan to integrate experimental workflow functionality into ELG.

### 3.5. Data Sets and Language Resources

The ELG consortium has defined an LR identification and sharing strategy. It starts by liaising with and capitalizing on existing activities to ingest LRs into the ELG. We have started by focusing on providers who are part of the consortium (ELDA/ELRA and ELG) and on recent activities such as ELRC-SHARE (Lösch et al., 2018; Piperidis et al., 2018) and META-SHARE (Piperidis, 2012; Piperidis et al., 2014). Table 2 provides an overview of what has been identified in these repositories and what is planned to be ingested into ELG, if their access and licensing conditions allow it.

|  | Corpora | Lexicons | Models | Total |
|---|---|---|---|---|
| ELRA | 848 | 1084 | 0 | 1932 |
| ELRC-SHARE | 396 | 132 | 0 | 528 |
| META-SHARE | 1580 | 1261 | 18 | 2859 |
| ELG | 78 | 109 | 76 | 263 |
| Total | 2902 | 2586 | 94 | 5582 |

Table 2: Identified LRs in the ELG consortium

LR modalities covered are text (corpora, lexicons, etc.), speech/audio, video/audiovisual, images/OCR, sign language, and others. About 220 additional repositories have been located so far, which will increase the numbers in Table 2 as the exploration and ingestion of LRs is progressing. Over 400 LRs from ELRA, ELRC-SHARE and META-SHARE have been selected for integration into ELG Release 1 (April 2020), subject to access conditions and license compliance, prioritizing open access resources. ELRA will provide over 100 LRs covering a wide range of modalities and languages. The languages addressed are either EU ones (e. g., Catalan, Czech, English, French, Frisian, German, Greek, Hungarian, Portuguese, Romanian, Russian, Serbian) or non-EU ones (e. g., Amharic, Arabic, Chinese, Hindi, Mongolian, Nepali, Pashto, Persian). The set also contains large multilingual resources. ELRC-SHARE currently offers over 200 resources, with a large number of TMX files and terminological databases. Many other resources are currently under clearing and cleaning. Around 100 LRs are integrated from META-SHARE. The remaining data (largely over 2000) will be checked and analysed.

As a first step and following the established metadata schema (Section 3.2), ELG has been working both on data integration procedures, where metadata compliance is key for the exchange of data and metadata descriptions, and on the implementation of features such as upload/download, licensing, billing, payment, etc. Metadata have been converted for the ELDA/ELRA, ELRC-SHARE and META-SHARE metadata schemas and a set of resources has already been ingested from each of them into ELG (Table 3).

|  | Corpora | Lexicons | Models |
|---|---|---|---|
| ELRA | 20 | 2 | – |
| ELRC-SHARE | 180 | 7 | – |
| META-SHARE | 52 | 12 | 7 |

Table 3: LRs available in ELG Release 1 alpha

### 3.6. Stakeholders and Community

ELG aims to respond to the challenge of Europe's fragmented European LT landscape (Vasiljevs et al., 2019), both with regard to industry and research. We address this issue by bringing together all stakeholders under a common umbrella platform, which is why outreach, communication and further community building play a crucial role in ELG. Our main target users are described in Section 3.6.1. In addition, we have been setting up two community instruments, the National Competence Centres (Section 3.6.2) and the European LT Council (Section 3.6.3).

#### 3.6.1. Key Stakeholders

ELG caters, first, for *commercial LT providers* who want to showcase their products, services and their own organisation. We want to provide *the* marketplace for European LT, which requires a broad geographical, technological and sectorial representation of, ideally, all relevant European provider companies. As with all stakeholder groups who will be represented in ELG, we initially populate the catalogue with records we take from existing databases, fully respecting GDPR. Once populated, representatives of an organisation will be able to claim (or delete) their record through the ELG user interface so that they can take over maintenance and populate their ELG page. To enable them to upload services and data sets, technical information, helper tools and also webinars and tutorials will be provided. The collaboration between ELG and the industry association LT Innovate helps us reach out to this stakeholder group to understand their demands and to make sure that their feedback finds its way back into the ELG. The more the platform meets the business requirements, the more likely LT providers will be to use and promote it and to use it as one additional or maybe even their preferred marketplace. *Research centres and universities* are also LT providers but their interest is not a monetary but a research-driven one. This stakeholder group provides data sets or smaller tools including rudimentary, experimental services that have evolved from research projects, rather than fully-fledged services that are ready for production and monetisation. For researchers, dissemination and further development of their tools and the exchange with other academics is the main driver to use the ELG. *LT users* are the most diverse target group. It includes organisations who want to make use of LT, students doing research for a paper, or job seekers. Members of this group can be on the lookout for information, try to find free services or be potential buyers. They interact with the ELG in the role of a consumer or potential customer. The six *ICT-29-2018 subtopic b) projects* are a special stakeholder group, as their consortia consist of research centres and universities as well as companies. The projects deal with domain-specific, challenge-oriented LT and provide services, tools and data sets which can also be showcased in the ELG. These projects can make use of the various functionalities as well as of the vast ELG community network. ELG is collaborating with all projects.[15] To establish ELG within the LT scene and to avoid silo-thinking, we collaborate with *all relevant existing projects and initiatives* in the field and also in all relevant neighbouring areas. ELG is collaborating or in the process of setting up collaborations with these related projects and initiatives that share this approach, such as AI4EU, ELRC, BDVA, CLAIRE, CLARIN, HumanE-AI, META-NET and various others (Rehm et al., 2020c). The *participants in the pilot projects* are also key stakeholders (see Section 3.7).

### 3.6.2. National Competence Centres

The ELG project set up 32 National Competence Centres (NCCs) to extend the reach of the ELG platform and initiative.[16] The NCCs were selected based on their involvement in relevant community initiatives. The fact that all NCCs have good connections to major local industry sectors while being part of academic organisations, guarantees independence from economic interests while ensuring sufficient outreach into commercial fields to serve the purpose of ELG. The NCCs function as bridges between the national and regional markets and the ELG, both as a platform and project. They provide information about stakeholders, services, data sets, resources and technologies from the given region. They know the language(s) and the political as well as economic situation in their countries and are represented in regional networks. The NCCs serve as multipliers when it comes to informing and promoting the ELG locally.

### 3.6.3. European LT Council

In addition to the NCCs, ELG is initiating a second new body, the European LT Council (LTC), as a pan-European group in which strategic LT-related matters can be discussed and coordinated. While the main purpose of the NCCs is to support the mission of the ELG project, the main goal of the LTC is to support and represent the European LT community. The LTC is meant to be a forum that enables easy and efficient communication and coordination at the European level, specifically with regard to ongoing and emerging international and also national activities relating to LT research, development and innovation. The LTC fosters the coordination and strategic as well as political discussion, representing all relevant stakeholder groups. It will prepare strategic recommendations, especially geared towards national and European administrations and funding agencies. Together with the NCCs, ELG has been assembling representatives from all important stakeholder groups relevant for the Multilingual Europe topic in the LTC for the purpose of establishing a platform and forum that enables a structured dialogue with all relevant stakeholders. The LTC's objectives are the discussion and coordination with regard to, among others, the following set of topics: Multilingual Europe and technology-enabled Multilingualism; Multilingual Digital Single Market; Language equality in the digital age; Digital language extinction; Technologies for lesser used or low resourced languages; Language-centric AI and its potentials and others. Particularly, the European LT Council discusses challenges, strategies, approaches, and solutions concerning the topics mentioned above; it coordinates and networks with national and international initiatives and organisations, and it drafts recommendations for national and international administrations and funding agencies.

### 3.7. Open Calls for Pilot Projects

ELG will provide close to 30% of its overall budget for a set of 15-20 small scale demonstrator pilot projects in the form of grants awarded after a call for proposals. The pilot projects will broaden ELG's portfolio by developing missing services or solutions that support underrepresented languages. At the same time, they will demonstrate the ELG's usefulness as a technology platform. The projects' results will be made available through the ELG. LT tools or services will be integrated into the ELG itself and made generally available. Applications using LT components will be included in the ELG catalogue.

The main objective of the open calls is to support SMEs that have long-term potential to either (a) contribute services, tools or data sets to the ELG to increase its coverage or (b) develop applications using LTs available in the ELG.

Thus, each applicant is allowed to submit up to two proposals, one for Objective (a) and one for Objective (b). Only SMEs and research organisations (including but not limited to higher education organisations, independent research organisations and NGOs) will be allowed to apply.

The selected projects will be supervised by the ELG Pilot Board, which consists of members of the ELG consortium. It provides a forum in which the ELG project partners can discuss the progress of the pilots, exchange feedback and monitor the results. It will be the technical and strategic interface between the pilot projects and ELG, where ELG can maximise its benefit from supporting the pilots and also ensure maximum benefit of the pilots with regard to the ELG. Financial support will be awarded to selected applicants following an open, transparent and expert-evaluation based selection process. Each proposal will be evaluated by three independent experts for the following criteria: (a) objective fit; (b) technical approach; (c) business, integration and dissemination plan; (d) budget adequacy; and (e) team. The first call was published in March 2020, the second will be published in September 2020. Both have a two months submission period. While the first call reflects the partial completion of the ELG, both calls share the same objectives and procedures. The selected projects will start in June 2020 and January 2021, respectively, their duration is expected to be in the 9-12 months range.

## 4. Sustainability through a Legal Entity

Achieving the intended scale of the ELG requires a high availability and performance of the overall system, service level agreements (SLAs) for the (paid) services as well as billing and support facilities. These characteristics create various non-trivial costs, that can only be covered ade-quately through a sustainable, long-term operational model. Costs include cloud hosting and bandwidth, personnel costs for operations, development, accounting, marketing, support and management, legal consulting (SLAs, GDPR, contracts etc.), office space, computers, electricity etc.

---
[15]https://www.european-language-grid.eu/meta-forum-2019/.
[16]https://www.european-language-grid.eu/ncc/

ELG is currently supported through one European project but meant to be a sustainable activity. To achieve this goal, we need to identify ways to cover the incurred costs on a long-term basis. For this purpose, we will establish a legal entity by approx. Q1/2021. Among the options are a for profit or non-profit company, a professional stakeholder association and a foundation, cf. (Hummel et al., 2016).

There are various potential ingredients of a future ELG business and operations plan. These include regular online ads (for companies, services, conferences etc.), sponsored content (e. g., first search result, clearly marked as "sponsored"), i. e., sponsored services, data sets, or companies, among others. The ELG legal entity can also offer training events, tutorials or webinars for a fee for commercial players, while keeping them free for academia. ELG conferences may include registration fees for delegates from industry, also offering sponsorship packages for companies. Furthermore, consulting services around ELG and language-centric AI can be offered. If we decide to establish a professional business association, membership fees could be part of the business plan. Furthermore, project grants can be used to sustain part of the operation. Additionally, the hosting of commercial LT services, models or data sets can be associated with a certain fee, while the results of publicly funded research can be made available for free, but the hosting costs would need to be covered nonetheless. In that regard, ELG could function as the secondary or maybe even primary dissemination channel for research projects or for companies that develop LT. Part of the ELG business model could also include the brokering of commercial LT services for a fee, with a split between the service owner and ELG as the broker. ELG could also function as a paid hoster for whole service or data repositories.

A survey demonstrates that the European LT community is very interested in the setup of the ELG (Melnika et al., 2019a). The results show that research and industry have a keen interest in depositing data sets and also code online. Many of the respondents use repositories for sharing tools, data sets and also annotated data sets, mentioning, as motivation, promotion, goodwill and reproducibility. However, there are also difficulties and challenges, including copyright and IPR, licenses and the submission process itself. The majority of the respondents is interested in providing functional services as containers and describes a central European LT platform as "important" or "very important".

## 5. Related Work

**Research Projects, Platforms, Initiatives** All in all, we have collected more than 30 projects, platforms and initiatives that are, in one way or another, relevant for ELG. Table 4 in the appendix shows a representative subset. They share at least one of the following goals with ELG: 1) they provide a collection of LT/NLP tools or data sets; 2) they provide a unified platform, which, underneath, harvests metadata records of data sets or services or tools from distributed sources; 3) they provide a sharing platform for the exchange of tools or data sets among stakeholders.

**Global Technology Enterprises** Many of the global technology enterprises offer a wide range of different processing services, beyond language, including cloud and compute resources, storage, different types of databases, data analytics, and also more engineering-related services such as encryption, development and deployment. Among these are offerings by Amazon, especially AWS[17] and Comprehend[18], Microsoft Azure Cognitive Services (Del Sole, 2018), the Google Cloud Platform[19] and the IBM Cloud (Kochut et al., 2011). Furthermore, Google has recently (Sept. 2018) released a dedicated search platform for data sets.[20]

## 6. Conclusions and Next Steps

It has repeatedly been argued that Europe should by no means outsource its multilingual communication and language challenge to providers from other continents since the European demands are so unique and complex (Rehm and Uszkoreit, 2013; Rehm, 2017; Rehm et al., 2020c). Instead, Europe should make use of its own excellent LT community. One of the obstacles to be overcome along the way is the creation of a shared platform for the whole community. The ELG will foster language technologies *for Europe* built *in Europe*, tailored to our languages and cultures and to our societal and economical demands, benefitting the European citizen, society, innovation and industry. There is currently no other scalable cloud platform that can play the role as a joint marketplace and broker for such a broad variety of services and data sets as we have foreseen for the ELG.

At the end of its first year, the three-year ELG project has already seen the first public demo of a fully functional miminum viable product of the ELG platform at META-FORUM 2019. Work in all three ELG areas is progressing at a fast pace. The last major milestones was the launch of the first open call in March 2020 and, at the same time, the launch of the first version (Release 1 alpha) of the ELG platform to interested parties. This version includes the first batches of functional services and data sets. The second open call will be published in September 2020, coinciding with Release 2 of the platform, services and data sets. Release 3 of the platform (including additional services and data sets) is foreseen for the last quarter of 2021. In 2020 and 2021 we will organise two more annual ELG conferences that will also include NCC and LTC meetings. At the end of 2021, a new legal entity will take over the further development and maintenance of the ELG platform. With regard to upcoming funding programmes on the European level, we foresee ELG to play a number of roles, especially as the main data and service provision and dissemination platform for the European LT and language-centric AI community (Rehm et al., 2020c) in Horizon Europe and Digital Europe Programme but also in national funding initiatives.

---

[17] https://aws.amazon.com
[18] https://aws.amazon.com/en/comprehend/
[19] https://cloud.google.com
[20] https://toolbox.google.com/datasetsearch

## Funding notice

The work presented in this paper has received funding from the European Union's Horizon 2020 research and innovation programme under grant agreement no. 825627 (European Language Grid) and from the German Federal Ministry of Education and Research (BMBF) through the project QURATOR (Wachstumskern no. 03WKDA1A).

## 7. Bibliographical References

# Appendix

| | Main NLP/LT Areas | Modalities | Languages Covered | Scaling | APIs & Interfaces | Environment | Goal, Scope, Ambition | Usability & Maintenance | Status | Costs | Market | Key Reference |
|---|---|---|---|---|---|---|---|---|---|---|---|---|
| European Language Grid (ELG) | Platform for data and services (see this article) | Text, speech | EU-24 plus many more minority and regional languages (and beyond) | Horizontally grid-able | GUI, REST | Cluster, cloud (Kubernetes) | Primary European platform for commercial and non-commercial LT | Customised, automated and intuitive GUI and documentation | Ongoing initiative (2019-2021) | n. a. | Europe | (this article) |
| Acumos AI[1] | Generic ML models as microservices | Text, speech | n. a. | Dockerized, cloud-ready | APIs | Cloud, local | Combination and integration of models, libraries, platforms | Well-documented, very complex | Ongoing initiative | Free | Global | (Zhao et al., 2018) |
| Alveo[2] | Platform for data and services | Speech, text, music, video | Various corpora and data sets | n. a. | GUI, R, Galaxy, NLTK API | Cluster, cloud | Scientists, researchers | Well-documented, faceted search | Project completed (2014–2018) | Free | Australia | (Cassidy et al., 2014) |
| Apache UIMA[3] | Standardized data mining framework | Text, speech | n. a. | Cluster support | Java, C++ APIs | Monolithic and grid (DUCC) | Focus on standardization of interfaces and modules | Programming skills required | Ongoing initiative | Free | Global | (Ferrucci and Lally, 2003) |
| CESSDA ERIC[4] | Platform for data and services | Various data formats | Primary focus on data (instead of languages) | In progress via SSHOC project | Catalogue GUI | ICA, vSphere | ERIC since 2017 | Catalogue search, documented website | Ongoing initiative | Free | Social science (Europe) | (Dekker, 2020) |
| CLARIN ERIC[5] | Platform for data and services | Text, speech | Various languages | Distributed infrastructure | GUI, APIs (tool-specific) | Distributed over several centers | ERIC since 2012 | Focus on catalogue functionality | Ongoing initiative | Free | SSH (Europe) | (Hinrichs and Krauwer, 2014) |
| DKPro[6] | Collection of approx. 20 NLP tools | Mainly text | Up to 70 languages, esp. EN, DE, FR, ES | n. a. | Tool-specific | Hosted on GitHub, Maven Central | Initiated by TU Darmstadt since 2007 | Mainly written in Java | Ongoing initiative | Free | Global | (Gurevych et al., 2007) |
| EOSC-hub[7] EOSC[8] | Catalogue of service, tool and resource providers | Cross-area publications and training materials | EU languages | Attempts to build a hub of horizontal infrastructures | Cloud-native | Many clouds and federated infrastructures | Hub of horizontally distributed computing resources | n. a. | Ongoing initiative (2018-2020) | Mixed | Europe | (EOSC-Governance-Board, 2018) |
| Language Grid[9] | Platform for data and services | Mainly text | EN, JA, others are tool-specific | n. a. | GUI | – | Anybody can contribute their own services | Manuals redirected to Sourceforge | Ongoing initiative (since 2017) | Small fee | Japan | (Murakami, 2017) |
| META-SHARE[10] ELRC-SHARE[11] | Platforms for data sets and LR | Mainly text (LRs, some LTs) | EU languages (and some others) | n. a. | GUI; harvesting via OAI-PMH | Distributed network | Network of repositories storing LRs | Different user roles | Ongoing initiative | Mixed | Europe | (Piperidis, 2012; Piperidis et al., 2014) (Piperidis et al., 2018) |
| OpenMinTeD[12] | Platform for content and services | Text, data | Language agnostic | Dockerized, cloud-ready | REST APIs, workflow editor | Platform | Research, Open Science, innovation | Easy access | Project completed (2015–2018) | Mixed | Europe | (Labropoulou et al., 2018) |
| Stanford CoreNLP API[13] | Integrated NLP toolkit (Java) | Text | EN, DE, FR, ES, AR, ZH | n. a. | REST APIs | Monolithic, single tools | High-quality NLP and LT tools | Basic programming skills required | Ongoing initiative | Free | Global | (Manning et al., 2014) |
| TextFlows[14] | Workflow platform for Text NLP and Text Mining | n. a. | n. a. | Horizontally, cluster support | GUI, editor for workflows | Cloud | Open source platform for NLP and TM workflows | Tutorials, test workflows | Project completed (2010–2016) | Free | Global | (Perovšek et al., 2016) |
| Weblicht[15] | Corpus annotation platform | Text (corpora) | EN, DE, FR, IT, HU etc. | n. a. | GUI, REST, OAI-PMH | Platform | Java developers can create services | Login via DFN-AAI or CLARIN account | Ongoing initiative | Free | Global | (Hinrichs et al., 2010) |

[1] Acumos AI: Making Artificial Intelligence Accessible to Everyone
[2] Alveo: Virtual Lab for Human Communication Science
[3] Apache UIMA: Open Source implementation of the Unstructured Information Management Applications OASIS specification
[4] CESSDA ERIC: Consortium of European Social Science Data Archives
[5] CLARIN ERIC: European Research Infrastructure for Language Resources and Technology
[6] DKPro: Community of projects focussing on re-usable Natural Language Processing software
[7] EOSC-hub: Services for the European Open Science Cloud
[8] EOSC: European Open Science Cloud
[9] Language Grid: Multilingual service platform which enables easy registration and sharing of language services
[10] META-SHARE: Open and secure network of repositories for sharing and exchanging language data and tools
[11] ELRC-SHARE: LR repository for language data and tools pertinent to Automated Translation (CEF eTranslation)
[12] OpenMinTeD: Open Mining Infrastructure for Text and Data
[13] Stanford CoreNLP API: Stanford CoreNLP provides a set of human language technology tools
[14] TextFlows: Open-source online platform for composition, execution, and sharing of interactive text mining and NLP workflows
[15] Weblicht: Execution environment for automatic annotation of text corpora

Table 4: European Language Grid and related research projects, platforms, initiatives (ELG is listed first, followed by the other projects, platforms, initiatives in alphabetical order)

|  | **ELG Partners** | | | | | **Non ELG** | | | | |
| --- | --- | --- | --- | --- | --- | --- | --- | --- | --- | --- |
|  | **A** | **B** | **C** | **D** | **Total** | **A** | **B** | **C** | **D** | **Total** |
| **ASR** | **12** | **3** | **9** |  | **24** | **8** | **1** | **2** | **2** | **13** |
| Speech Recognition | 12 | 3 | 9 |  | 24 | 8 | 1 | 2 | 2 | 13 |
| **IE & Text Analysis** | **311** | **50** | **119** | **12** | **492** | **304** | **114** | **144** | **429** | **991** |
| Terminology search | 24 | 1 | 1 |  | 26 |  |  |  |  | 0 |
| Term recognition | 24 | 1 | 1 |  | 26 |  |  |  |  | 0 |
| Part of speech tagging | 24 | 7 | 13 |  | 44 | 22 | 8 | 11 | 16 | 57 |
| Dependency parsing | 24 | 7 | 13 |  | 44 | 22 | 7 | 11 | 11 | 51 |
| Terminology markup | 24 | 1 | 1 |  | 26 |  |  |  |  | 0 |
| Lemmatisation | 24 | 7 | 12 |  | 43 | 22 | 7 | 11 | 9 | 49 |
| Morphological analyser | 24 | 7 | 13 |  | 44 | 23 | 11 | 16 | 89 | 139 |
| Tokenization | 24 | 7 | 13 |  | 44 | 23 | 10 | 12 | 19 | 64 |
| Language identification | 22 | 4 | 14 | 12 | 52 | 24 | 11 | 16 | 136 | 187 |
| Named entity recognition | 18 | 5 | 11 |  | 34 | 21 | 8 | 10 | 5 | 44 |
| Keyword extraction | 9 | 3 | 9 |  | 21 |  |  |  |  | 0 |
| Sentiment analysis | 8 |  | 4 |  | 12 | 23 | 11 | 16 | 91 | 141 |
| Summarization | 7 |  | 5 |  | 12 |  |  |  |  | 0 |
| Key phrase extraction | 7 |  | 5 |  | 12 |  |  |  |  | 0 |
| Polarity detection | 7 |  | 4 |  | 11 |  |  |  |  | 0 |
| Categorization | 5 |  |  |  | 5 |  |  |  |  | 0 |
| Sentence splitting | 4 |  |  |  | 4 | 10 | 4 | 3 | 3 | 20 |
| NER disambiguation | 4 |  |  |  | 4 |  |  |  |  | 0 |
| Information extraction | 3 |  |  |  | 3 |  |  |  |  | 0 |
| Textual entailment | 3 |  |  |  | 3 | 1 |  |  |  | 1 |
| Number annotation | 2 |  |  |  | 2 | 6 | 2 | 1 |  | 9 |
| Time annotation | 2 |  |  |  | 2 |  |  |  |  | 0 |
| Date detection | 2 |  |  |  | 2 | 5 | 2 | 1 |  | 8 |
| Proofing tools | 2 |  |  |  | 2 | 1 |  |  |  | 1 |
| Entity linking | 2 |  |  |  | 2 |  |  |  |  | 0 |
| Text extraction | 1 |  |  |  | 1 |  |  |  |  | 0 |
| Measurement annotation | 1 |  |  |  | 1 |  |  |  |  | 0 |
| Negation detection | 1 |  |  |  | 1 |  |  |  |  | 0 |
| Relation extraction | 1 |  |  |  | 1 |  |  |  |  | 0 |
| Measurement normalisation | 1 |  |  |  | 1 |  |  |  |  | 0 |
| Opinion mining | 1 |  |  |  | 1 |  |  |  |  | 0 |
| Parsing | 1 |  |  |  | 1 | 6 | 1 | 1 |  | 8 |
| Noun phrase extraction | 1 |  |  |  | 1 | 1 |  |  |  | 1 |
| Number normalisation | 1 |  |  |  | 1 |  |  |  |  | 0 |
| Intent extraction |  |  |  |  | 0 |  |  |  |  | 0 |
| Word segmentation |  |  |  |  | 0 | 1 |  | 1 |  | 2 |
| Multiword detection |  |  |  |  | 0 |  |  |  | 1 | 1 |
| Prepositional phrase attachment |  |  |  |  | 0 | 1 |  |  |  | 1 |
| Tagging |  |  |  |  | 0 | 1 |  | 1 |  | 2 |
| Mention detection |  |  |  |  | 0 | 1 |  | 1 |  | 2 |
| Quantity detection |  |  |  |  | 0 | 3 | 2 | 1 |  | 6 |
| Transliteration |  |  |  |  | 0 |  |  |  | 1 | 1 |
| Language modeling |  |  |  |  | 0 |  |  |  |  | 0 |
| Coreference resolution |  |  |  |  | 0 | 2 |  | 1 |  | 3 |
| Relationship extraction |  |  |  |  | 0 | 1 |  |  |  | 1 |
| Collocation extraction |  |  |  |  | 0 | 1 |  |  |  | 1 |
| Semantic reasoning |  |  |  |  | 0 | 1 |  |  |  | 1 |
| Anaphora resolution |  |  |  |  | 0 | 1 |  |  |  | 1 |
| Semantic role labeling |  |  |  |  | 0 | 3 | 1 |  |  | 4 |
| Constituency parsing |  |  |  |  | 0 | 4 |  | 2 |  | 6 |
| N-grams |  |  |  |  | 0 | 5 | 3 |  |  | 8 |
| Phonetic encoding |  |  |  |  | 0 | 2 |  |  |  | 2 |
| Idiom extraction |  |  |  |  | 0 | 1 |  |  |  | 1 |
| Word frequencies |  |  |  |  | 0 | 23 | 10 | 13 | 23 | 69 |
| Shallow parsing |  |  |  |  | 0 | 3 | 3 |  | 1 | 7 |
| Word sense disambiguation |  |  |  |  | 0 | 4 | 2 |  | 1 | 7 |
| Stemming |  |  |  |  | 0 | 12 | 1 | 2 |  | 15 |

|  | ELG Partners | | | | | Non ELG | | | | |
| --- | --- | --- | --- | --- | --- | --- | --- | --- | --- | --- |
|  | A | B | C | D | Total | A | B | C | D | Total |
| Noun phrase frequencies |  |  |  |  | 0 | 23 | 10 | 13 | 23 | 69 |
| Open information extraction |  |  |  |  | 0 | 1 |  |  |  | 1 |
| **MT** (Source ↓ / Target →) | **67** | **2** | **7** | **1** | **77** | **44** | **17** | **6** | **25** | **92** |
| A | 58 | 2 | 7 | 1 | 68 | 14 | 10 | 2 | 12 | 38 |
| B | 2 |  |  |  | 2 | 16 | 2 |  | 5 | 23 |
| C | 7 |  |  |  | 7 | 2 | 1 | 2 | 2 | 7 |
| D |  |  |  |  | 0 | 12 | 4 | 2 | 6 | 24 |
| **Text to Speech** | **7** | **1** | **1** | **2** | **11** |  |  |  |  | **0** |
| Text to speech | 7 | 1 | 1 | 2 | 11 |  |  |  |  | 0 |
| **Other** | **27** | **2** |  |  | **29** |  |  |  |  | **0** |
| Pipeline | 1 |  |  |  | 1 |  |  |  |  | 0 |
| Text indexing | 2 |  |  |  | 2 |  |  |  |  | 0 |
| Web crawler | 24 | 2 |  |  | 26 |  |  |  |  | 0 |
| **Total general** | **424** | **58** | **136** | **15** | **633** | **356** | **132** | **152** | **456** | **1096** |

Table 5: Number of services per language category provided by ELG partners and other potential providers out of the consortium

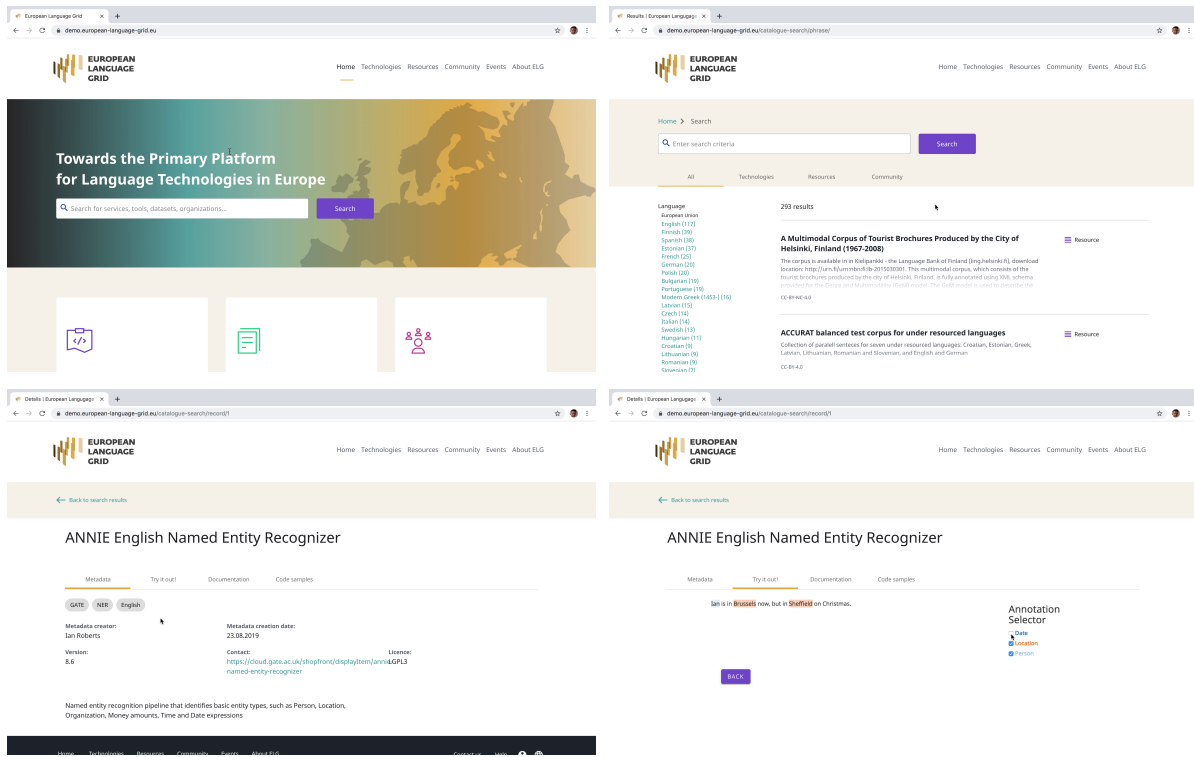

Figure 3: The interactive ELG user interface (first minimum viable product version, October 2019)

| Source Language | | Target Language | | |
|---|---|---|---|---|
| Category | Language | Category | Language | Services |
| **A** | **Bulgarian** | A | English | 2 |
| | **Czech** | A | English | 2 |
| | **Danish** | A | English | 2 |
| | **English** | C | Arabic | 2 |
| | | A | Bulgarian | 2 |
| | | C | Chinese | 2 |
| | | A | Czech | 2 |
| | | A | Danish | 2 |
| | | A | Estonian | 3 |
| | | A | Finnish | 2 |
| | | A | French | 2 |
| | | A | German | 2 |
| | | D | Hindi | 1 |
| | | A | Latvian | 3 |
| | | A | Lithuanian | 2 |
| | | B | Norwegian | 2 |
| | | A | Polish | 2 |
| | | A | Portuguese | 1 |
| | | A | Romanian | 1 |
| | | C | Russian | 3 |
| | | A | Spanish | 2 |
| | | A | Swedish | 2 |
| | **Estonian** | A | English | 3 |
| | **Finnish** | A | English | 2 |
| | **French** | A | English | 2 |
| | **German** | A | English | 4 |
| | **Latvian** | A | English | 3 |
| | **Lithuanian** | A | English | 2 |
| | **Polish** | A | English | 2 |
| | **Portuguese** | A | English | 1 |
| | **Romanian** | A | English | 1 |
| | **Spanish** | A | English | 2 |
| | **Swedish** | A | English | 2 |
| **Total A** | | | | **68** |
| **B** | **Norwegian** | A | English | 2 |
| **Total B** | | | | **2** |
| **C** | **Arabic** | A | English | 2 |
| | **Chinese** | A | English | 2 |
| | **Russian** | A | English | 3 |
| **Total C** | | | | **7** |
| **Total general** | | | | **77** |

Table 6: Language pairs supported by machine translation services provided by ELG partners